\documentclass[acmtog,screen,review=false]{acmart}
\AtBeginDocument{%
  }

\usepackage{etoolbox,lipsum}
\usepackage{multirow}
\usepackage{tabularx}
\newcolumntype{Y}{>{\centering\arraybackslash}X}

\acmJournal{TOG}
\begin{document}

\title{DMF-Net: Image-Guided Point Cloud Completion with Dual-Channel Modality Fusion and Shape-Aware Upsampling Transformer}

\author{Aihua Mao}
\authornote{Both authors contributed equally to this research.}
\email{ahmao@scut.edu.cn}
\author{Yuxuan Tang}
\authornotemark[1]
\email{202120144052@mail.scut.edu.cn}
\affiliation{
  \institution{South China University of Technology}
  \country{China}
}
\author{Jiangtao Huang}
\affiliation{
  \institution{South China University of Technology}
  \country{China}
}
\author{Ying He}
\affiliation{
  \institution{Nanyang Technological University}
  \country{Singapore}}
\email{yhe@ntu.edu.sg}

\begin{abstract}
  In this paper we study the task of a single-view image-guided point cloud completion. Existing methods have got promising results by fusing the information of image into point cloud explicitly or implicitly. However, given that the image has global shape information and the partial point cloud has rich local details, We believe that both modalities need to be given equal attention when performing modality fusion. To this end, we propose a novel dual-channel modality fusion network for image-guided point cloud completion(named DMF-Net), in a coarse-to-fine manner. In the first stage, DMF-Net takes a partial point cloud and corresponding image as input to recover a coarse point cloud. In the second stage, the coarse point cloud will be upsampled twice with shape-aware upsampling transformer to get the dense and complete point cloud. Extensive quantitative and qualitative experimental results show that DMF-Net outperforms the state-of-the-art unimodal and multimodal point cloud completion works on ShapeNet-ViPC dataset.
\end{abstract}

\begin{teaserfigure}
    \centering
    \includegraphics[width=\textwidth]{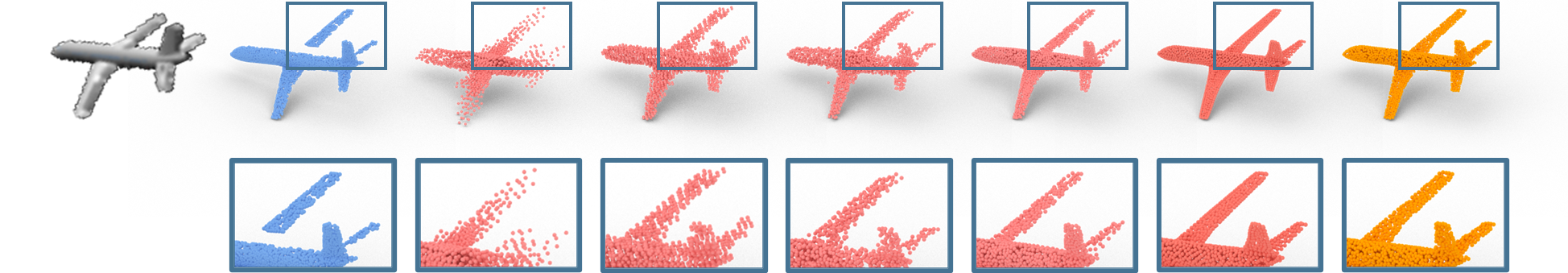}
    \begin{tabularx}{0.95\textwidth}{@{}Y@{}Y@{}Y@{}Y@{}Y@{}Y@{}Y@{}Y@{}}
    Image \ \ \ &
    Partial \ \ \ &
    PCN \ \ \ &
    PointAttN \ \ \ &
    ViPC \ \ &
    XMFnet \ \ \ &
    DMF-Net \ \ \ \ &
    GT \ \ \ \
    \end{tabularx}
    \caption{The visualization of point cloud completion results produced by PCN~\cite{yuan2018pcn}, PointAttN~\cite{wang2024pointattn}, ViPC~\cite{zhang2021view}, XMFnet~\cite{aiello2022cross} and DMF-Net. Given a partial point cloud and a corresponding single-view image, our DMF-Net reconstructs a high-quality complete point cloud with more local details and better uniformity.}
    \Description{A teaser image of DMF-Net.}
    \label{fig:teaser}
\end{teaserfigure}




\maketitle

\section{Introduction}

With the development of 3D sensing equipment such as LiDARs, laser scanners and RGB-D cameras, it is easier to acquire point clouds data, which has abundant applications like autonomous driving~\cite{cui2021deep}, scene understanding~\cite{hou20193d}, robotic vision~\cite{varley2017shape} and augmented reality~\cite{park2008multiple}. However, point clouds captured by these devices are often incomplete and sparse due to self-occlusions, occlusions between objects, uneven illumination and low scanning resolution. The poor quality point clouds can make it difficult to understand 3D shape, which will leads to ambiguity in many downstream tasks like point cloud detection~\cite{zhou2018voxelnet,qi2019deep}, point cloud reconstruction~\cite{yang2018foldingnet,mandikal2019dense} and point cloud upsampling~\cite{yu2018pu,mao2022pu}. Therefore, point cloud completion, which focus on recovering complete high quality point clouds from sparse partial input, has become a hot research topic.

Inspired by the seminal work~\cite{qi2017pointnet} employing shared-MLP for point-wise feature extraction, an increasing number of learning-based approaches for point cloud completion have emerged. Most of these learning-based works utilize an encoder-decoder architecture, wherein the encoder extracts a global latent vector from the partial input, and the decoder subsequently generates the complete point cloud based on this global latent representation. However, the global information from an incomplete point cloud can be ambiguous and misleading.

Compared to 3D point cloud data, it is obvious that the corresponding 2D image data is easier to acquire~\cite{nan20142d}. Moreover, we know that humans can easily infer the 3D shape of an object from its 2D image, which intuitively proves that 2D image can play an important role in 3D shape completion task. Specifically, an image has the global shape information of an object, while the partial point cloud has rich local details. The two complementary information mentioned before can make the extracted feature more representative so that the decoder can generate a more plausible shape. Previous multimodal completion networks~\cite{zhang2021view,zhu2023csdn,aiello2022cross} have successfully improved the performance by introducing a single-view image into the point cloud completion process. However, the modality fusion process of these methods is dominated by point cloud modality, lacking reasonable utilization of image modality.

To address the above issue, we design an image-guided point cloud completion network with dual-channel modality fusion and shape-aware upsampling transformer (named DMF-Net). DMF-Net recovers the partial point cloud in two stages. At the first stage, DMF-Net employs an encoder-decoder architecture to generate a sparse but complete point cloud. We observe that image modality should not just serve as a guidance for point cloud modality, but should complement point cloud modality, which indicates the two modalities should be considerd equally important when performing modality fusion. Thus, in the encoding phase, we propose a dual-channel modality fusion strategy, which is capable of capturing the complementary information in both image and point cloud by fusing the two modalities in a symmetric way. At the second stage, inspired by these works~\cite{xiang2021snowflakenet,zhou2022seedformer,wang2024pointattn}, to further recover the local details and improve the uniformity of the complete point cloud, based on the coarse point cloud generated by the first stage, DMF-Net utilizes a shape-aware upsampler to recover a dense and complete point cloud. The term shape-aware means during the upsampling process, the local feature context is utilized for better recovering the local geometric details.

In summary, the main contributions of this work are threefold:
\begin{itemize}
\item{We propose a novel point cloud completion network with dual-channel modality fusion and shape-aware upsampling transformer, which utilizes the complementary information in image and partial point cloud to recover a high quality complete point cloud in a coarse-to-fine manner.}

\item{We propose a dual-channel modality fusion module, which fuses image modality and point cloud modality in a symmetric way, allowing the two modalities contribute equally to the fused global feature.}

\item{We propose a novel shape-aware upsampling transformer for point cloud completion, which is capable of capturing the local geometric details through encouraging the communication of local neighborhood points with self-attention mechanism.}

\end{itemize}

\section{Related Work}

\subsection{Traditional Point Cloud Completion}
Traditional methods for point cloud completion can be roughly categorize into two types: geometry-based and alignment-based. Geometry-based methods usually utilize the visible part to predict the missing part by some prior geometric assumptions. Methods like~\cite{mitra2006partial,mitra2013symmetry,sung2015data} employ symmetric axes to repeat the regular part of objects from the observed regions. Other methods such as~\cite{davis2002filling,berger2014state,nguyen2016field} fill holes of a smooth surface by local interpolations. These works can only handle partial scans with a small degree of incompleteness like small holes on the surface. Alignment-based methods involve matching partial shapes with template models from a large database. Some methods~\cite{pauly2005example,shao2012interactive,li2015database} retrieve the whole 3D shapes directly. Other methods~\cite{kalogerakis2012probabilistic,shen2012structure,kim2013learning} retrieve part of the 3D shapes to do the completion task. These methods can handle variant incompleteness of point cloud, but they are not suitable for practical application scenarios due to the expensive cost of inference optimization and building large datasets, as well as their high sensitivity to noise.

\begin{figure*}[t]
  \centering
  \includegraphics[width=\textwidth]{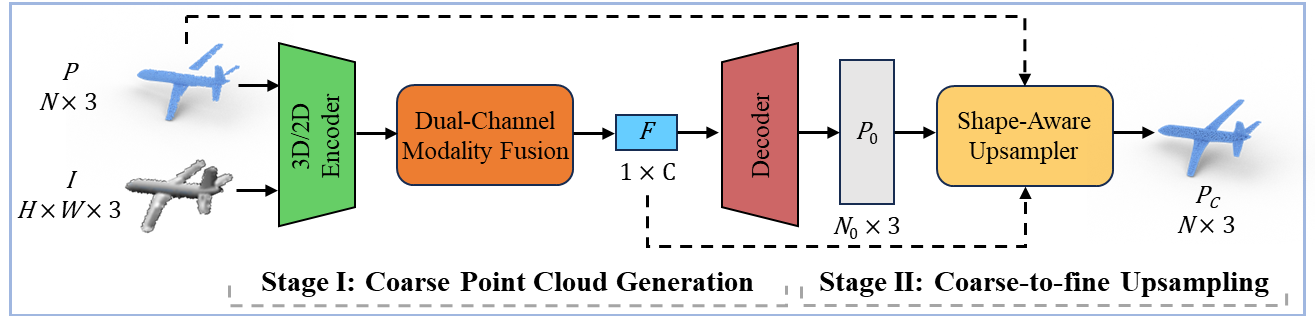}
  \caption{The architecture of our proposed DMF-Net. It takes two stages to recover the complete point cloud. In the first stage, a coarse point cloud is generated according to the partial point cloud and corresponding single-view image. In the second stage, the coarse point cloud will be upsampled to get the dense and complete output.}
  \Description{The architecture of DMF-Net.}
  \label{fig:pipeline}
\end{figure*}

\subsection{Unimodal Point Cloud Completion}

In recent years, deep learning-based methods have become the mainstream direction of point cloud completion. Early learning-based methods~\cite{dai2017shape,han2017high} apply 3D CNN to voxel-based representations of 3D objects, which are computational expensive. Thanks to PointNet~\cite{qi2017pointnet} and PointNet++~\cite{qi2017pointnet++}, most methods utilize an encoder-decoder architecture, where the encoder part extracts a global latent vector of the partial input and the decoder part reconstructs the complete point cloud through the global vector. PCN~\cite{yuan2018pcn} is the pioneering work which first avoids any geometric priors and annotations about the underlying shape. In the decoding phase, PCN employs the folding operation proposed by FoldingNet~\cite{yang2018foldingnet} to achieve point cloud refinement and upsampling. AtlasNet~\cite{groueix2018papier} represents a 3D shape as a collection of parametric surface elements and tackles the point cloud completion by learning the transformation of 2D square to 3D surface. TopNet~\cite{tchapmi2019topnet} proposes a hierarchical rooted tree structure decoder which can generate structured point clouds iteratively. MSN~\cite{liu2020morphing} introduces a coarse-to-fine method which generates different parts of a shape and refines the coarse point cloud with a residual network. GRNet~\cite{xie2020grnet} proposes a technique to convert the point cloud to 3D grid so that 3D CNN can be applied. PF-Net~\cite{huang2020pf} proposes a multi-scale encoder to extract both global and local features and a point pyramid decoder to generate points in missing regions hierarchically. Inspired by GAN~\cite{goodfellow2014generative}, adversarial loss is utilized in the training process to get better completion performance. CRN~\cite{wang2020cascaded} proposes a cascaded refinement network to iteratively upsample and refine the coarse point cloud. ECG~\cite{pan2020ecg} utilizes an encoder-decoder architecture to first restore the skeleton of a complete point cloud, then constructs a graph and applies convolution for edge perception, while expanding the point cloud features to achieve upsampling and refinement. VRC-Net~\cite{pan2021variational} reconstructs a coarse point cloud by employing a dual-path probabilistic modeling network and do the refinement by employing a relational enhancement network which can learn multi-scale structural relations between partial input and the coarse point cloud. PMP-Net~\cite{wen2021pmp} treats the point cloud completion as a deformation process that minimizes the total point moving distance, thereby enhancing the quality of the complete point cloud. Works like~\cite{yu2021pointr,xiang2021snowflakenet,zhou2022seedformer,wang2024pointattn} have successfully integrated the Transformer~\cite{vaswani2017attention} architecture into point cloud completion, further improve the performance of the completion network. PoinTr~\cite{yu2021pointr} reformulates the point cloud completion task into a set-to-set translation task, and introduces a geometric perception module to better capture the spatial neighborhood information of the point cloud. SnowflakeNet~\cite{xiang2021snowflakenet} regards the generation of a complete point cloud as a snowflake-like growth process of points in 3D space, and utilizes skip-Transformer to learn the point splitting mode that best adapts to the local area to generate a locally compact and structured complete point cloud. Seedformer~\cite{zhou2022seedformer} introduces local seed shape representation and integrates the spatial and semantic relationships between neighborhood points through Transformer during upsampling. PointAttN~\cite{wang2024pointattn} utilizes attention mechanism to achieve geometric detail perception and feature enhancement, 
and establish the structural relationship between points to generate a complete point cloud with detailed geometric structure.

However, the above methods only takes partial point cloud as input. When performing feature extraction on the partial input, the outcoming global latent vector actually does not contain the information of the missing part, which means the encoding process suffers from information loss that lowers the performance of the completion network.

\subsection{Multimodal Point Cloud Completion}

To solve the lack of global shape information in the partial point cloud, researchers have introduced image modality into the point cloud completion task, which is pioneered by ViPC~\cite{zhang2021view}. Extra modalities can provide complementary information which can be used to generated more plausible 3D shapes, and inspired by this idea, ViPC~\cite{zhang2021view} utilizes a pre-trained neural network to generate a point cloud according to the corresponding image, and then explicitly fuse the two modalities through concatenation between the partial point cloud and the generated point cloud. In the refinement stage, a point-wise offset vector is predict by utilizing the information of the two modalities. XMFnet~\cite{aiello2022cross}, Unlike ViPC, performs feature-level fusion to implicitly fuse the two modalities through cross-attention and self-attention, and employs multiple independent branches to predict different parts of the complete point cloud. Inspried by StyleGAN~\cite{karras2019style}, CSDN~\cite{zhu2023csdn} treats the modality fusion task as a style transfer task, and transfer the style of the image to the partial point cloud in the folding-based decoding process. CSDN proposes a dual-refinement module to predict a offset vector by leveraging both the global information in image and local information in partial input. These methods got promising results by utilizing image modality to provide the information of the missing part, however, the fusion process is dominated by the point cloud modality, and the uniformity of the final output is less satisfied.

\section{Method}

\subsection{Overview}
The overall framework of our proposed DMF-Net is shown in Fig.~\ref{fig:pipeline}, in which the cross-modal point cloud completion task consists of two stages: coarse point cloud generation and coarse-to-fine upsampling. Specifically, given the partial point cloud $P \in \mathbb{R}^{N \times 3}$, and the corresponding single-view image $I \in \mathbb{R}^{H \times W \times 3}$, our goal is to generate the complete point cloud $P_C \in \mathbb{R}^{N \times 3}$. $N$ denotes the number of points in $P$ and $P_C$. $H$ and $W$ denote the height and width of $I$. In the first stage, we adopt a 3D encoder and a 2D encoder to extract point cloud features and image features respectively. Then, we fuse the feature of the two modalities through the proposed dual-channel modality fusion module, which yields a global vector. Next, the decoder is utilized to generate a coarse point cloud according to the global vector. In the second stage, the coarse point cloud is upsampled by the shape-aware upsampler to get the complete point cloud $P_C$.

\subsection{Point Cloud and Image Encoders}

\begin{figure}[htbp]
  \centering
  \includegraphics[width=\linewidth]{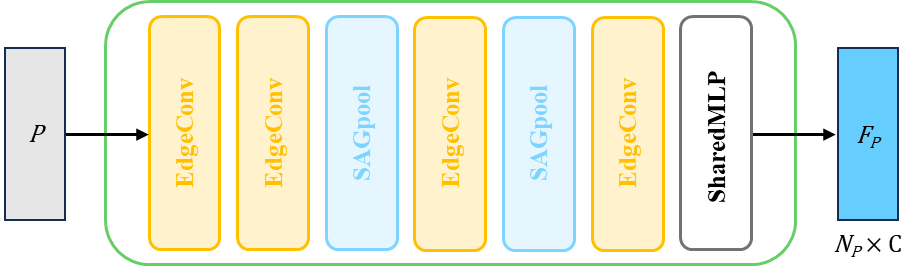}
  \caption{The architecture of 3D encoder, which is employed to extract the feature of partial point cloud.}
  \Description{The architecture of 3D encoder.}
  \label{fig:3d_encoder}
\end{figure}

\noindent{{\bf 3D Encoder.}}
The partial point cloud $P$ contains rich local details. In order to capture the local 
geometric information of $P$, we feed the partial point cloud into the 3D encoder, as shown in Fig.~\ref{fig:3d_encoder}. Inspired by~\cite{aiello2022cross}, the architecture of 3D encoder is a sequence of graph-convolutional layers (EdgeConv~\cite{phan2018dgcnn}) interleaved by graph-pooling layers (SAGPool~\cite{lee2019self}). The EdgeConv layer employs KNN algorithm to build graph of the partial point cloud and MLPs to extract point-wise feature of the graph. Directly use maxpooling operation on the point-wise feature will inevitably suffer from information loss, so the SAGPool layer is utilized for the purpose of reducing the number of points while preserving more local information. The last layer of the 3D encoder is a shared MLP, which is utilized to yield the point-wise feature $F_P \in \mathbb{R}^{N_P \times C}$. $N_P$ is the number of points and $C$ is the dimensions of the point-wise feature $F_P$.

\noindent{{\bf 2D Encoder.}}
The single-view image $I$ contains the global shape information which serves as a guiding information during the point cloud completion process. The simple but effective convolutional network ResNet18~\cite{he2016deep} is chosen as our 2D encoder. The extracted $7 \times 7 \times C$ feature map is reshaped to get the pixel-wise feature $F_I \in \mathbb{R}^{N_I \times C}$, where $N_I$ is the number of pixels and $C$ is the dimensions of $F_I$.

\subsection{Dual-channel Modality Fusion}

\begin{figure}[htbp]
  \centering
  \includegraphics[width=\linewidth]{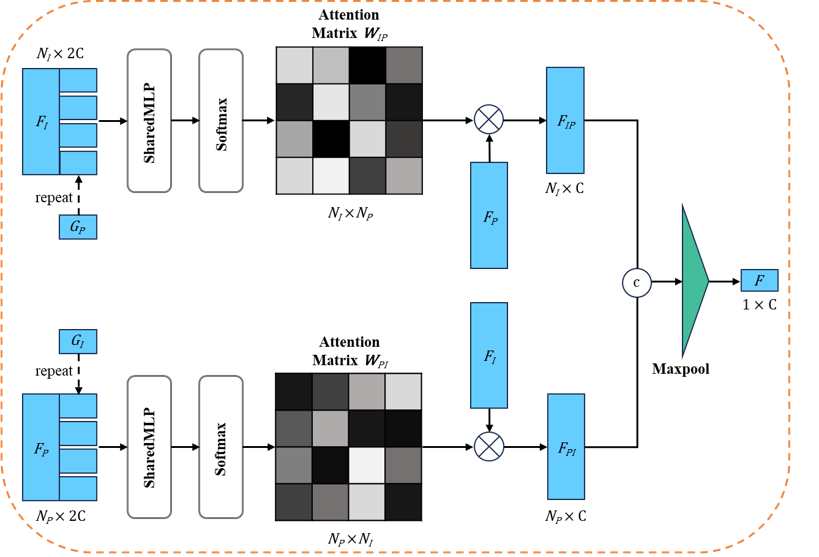}
  \caption{The architecture of Dual-channel Modality Fusion (DMF) module, which fuses the point-wise feature and the pixel-wise feature in a symmetric way.}
  \Description{The architecture of DMF module.}
  \label{fig:DMF}
\end{figure}

To better leverage the complementary information of the two modalities, we design a Dual-channel Modality Fusion (DMF) module, as shown in Fig.~\ref{fig:DMF}. Specifically, after getting the point-wise feature $F_P$ and the pixel-wise feature $F_I$, we utilize maxpooling operation on $F_P$ to get the point cloud global feature $G_P \in \mathbb{R}^{1 \times C}$ and $F_I$ to get the image global feature $G_I \in \mathbb{R}^{1 \times C}$. The upper path aims to calculate point cloud feature enhanced by image feature $F_{IP} \in \mathbb{R}^{N_{I} \times C}$, where $G_P$ is replicated for $N_I$ times and concatenate with $F_I$. Then, the attention matrix $W_{IP} \in \mathbb{R}^{N_{I} \times N_{P}}$ is calculated by

\begin{displaymath}
W_{IP}=\mathit{softmax}\left(\mu\left(F_{1}\right)\right),
\end{displaymath}

where $F_{1}=\left[F_{I}, \mathit{replicate}\left(G_{P}, N_{I}\right)\right]$ is the enhanced pixel-wise feature and $replicate(\cdot,\cdot)$ denotes the replication operation, $[\cdot,\cdot]$ denotes the concatenation operation and $\mu$ is a non-linear function implemented by shared-MLPs. Next, $F_{IP}$ is calulated by

\begin{displaymath}
F_{IP}=W_{IP} \otimes F_{P},
\end{displaymath}

where $\otimes$ denotes the matrix multiplication operation. Similarly, the lower path aims to calculate image feature enhanced by point cloud feature $F_{PI} \in \mathbb{R}^{N_{P} \times C}$. Attention matrix $W_{PI} \in \mathbb{R}^{N_{P} \times N_{I}}$ is calculated by

\begin{displaymath}
W_{PI}=\mathit{softmax}\left(\theta\left(F_{2}\right)\right),
\end{displaymath}

where $F_{2}=\left[F_{P}, \mathit{replicate}\left(G_{I}, N_{P}\right)\right]$ is the enhanced point-wise feature and $\theta$ is a non-linear function implemented by shared-MLPs. Then, $F_{PI}$ is calulated by

\begin{displaymath}
F_{PI}=W_{PI} \otimes F_{I},
\end{displaymath}

The enhanced features $F_{IP}$ and $F_{PI}$ are concatenated and maxpooling is utilized to yield the fused global feature $F \in \mathbb{R}^{1 \times C}$. DMF module fuses the feature of the two modalities in a symmetric way, and by doing so, the image modality can serve as a better guidance, which can enhance the representational ability of the fused global feature $F$.

\subsection{Coarse Point Cloud Generation}

\begin{figure}[htbp]
  \centering
  \includegraphics[width=\linewidth]{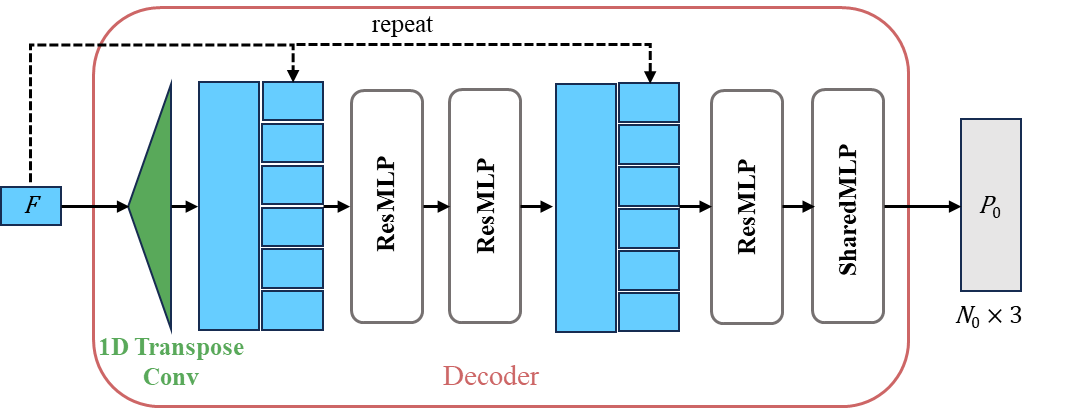}
  \caption{The architecture of decoder, which takes in the fused global feature and reconstructs a coarse point cloud.}
  \Description{The architecture of decoder.}
  \label{fig:decoder}
\end{figure}

To reconstruct a coarse point cloud $P_0 \in \mathbb{R}^{N_0 \times 3}$ from the fused feature $F$ (where $N_0$ is the number of points in $P_0$), we utilize the seed generator of~\cite{xiang2021snowflakenet} as our decoder, as shown in Fig.~\ref{fig:decoder}. The decoder first employs 1D transpose convolution to expand the global feature to a point-wise feature. Then concatenates it with duplicated global feature to preserve more information. Next, two residual shared MLPs are utilized to enhance the point-wise feature. The last layer of the decoder is a shared MLP to adjust the feature dimension to 3, thus outputs the coordinates of the coarse point $P_0$.

\subsection{Coarse-to-Fine Upsampling}

\begin{figure}[htbp]
  \centering
  \includegraphics[width=\linewidth]{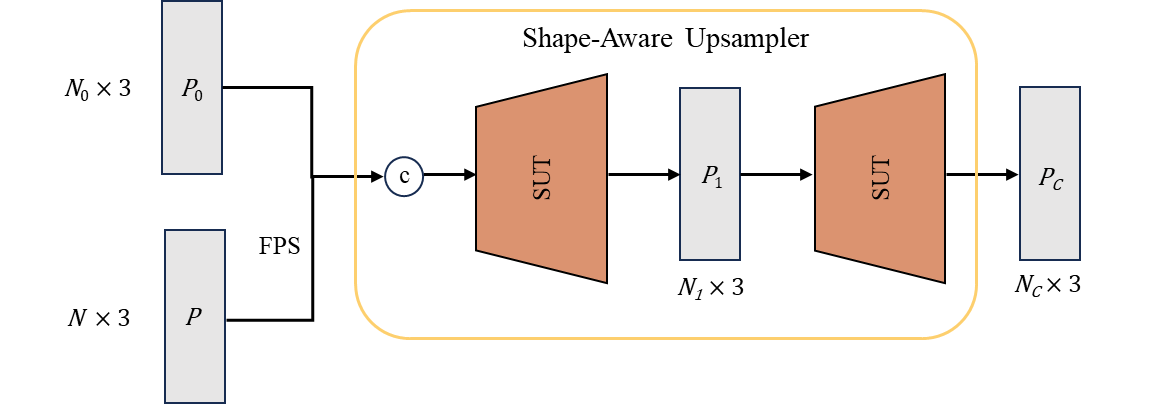}
  \caption{The architecture of upsampler, which consists of two consecutive shape-aware Upsample Transformers (SUTs) to generate the complete point cloud in a coarse-to-fine manner.}
  \Description{The architecture of upsampler.}
  \label{fig:upsampler}
\end{figure}

\begin{figure}[ht]
  \centering
  \includegraphics[width=\linewidth]{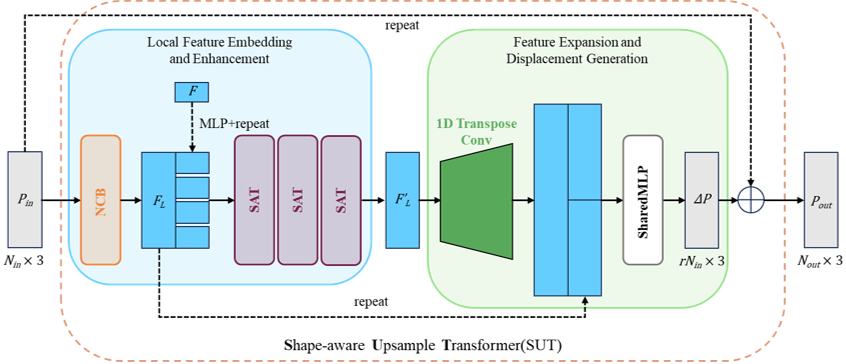}
  \caption{The architecture of SUT, which upsamples the coarse point cloud in two steps: 1) local feature embedding and enhancement, 2) feature expansion and displacement generation.}
  \Description{The architecture of SUT.}
  \label{fig:SUT}
\end{figure}

Existing multimodal completion networks like ViPC~\cite{zhang2021view} and CSDN~\cite{zhu2023csdn} both have adopted a coarse-to-fine completion strategy. Moreover, in the refinement stage, both of them leverage information of the two modalities to predict a displacement vector and add to the coarse point cloud to generate a point-wise displacement to achieve the refinement goal. However, the generated complete point cloud lacks local details due to the ignorance of the 
rich neighborhood information. Besides, the uniformity of the complete point cloud is less satisfied since the coarse point cloud and the complete point cloud have same points which means the refinement can only handle some noisy points. The coarse point cloud generated by the first stage is sparse, so we design a shape-aware upsampler to increase the number of points to get the dense and complete point cloud $P_C \in \mathbb{R}^{N_C \times 3}$, as shown in Fig.~\ref{fig:upsampler}. In order to preserve the local information in parital point cloud $P$, before upsampling, $P$ is downsampled by the FPS algorithm and concatenated with the coarse point cloud $P_0$. Then, intermediate point cloud $P_1 \in \mathbb{R}^{N_1 \times 3}$ is obtained through one shape-aware upsample Transformer (SUT). Another SUT is utilized to obatain the complete point cloud $P_C$. The architecture of SUT is illustrated in Fig.~\ref{fig:SUT}.

\noindent{{\bf Local Feature Embedding and Enhancement.}}
To make use of the rich neighborhood information during the upsampling process, given a sparse point cloud $P_{in} = \{p_i\}_{i=1}^{N_{in}}$, where $N_{in}$ is the number of points in $P_{in}$ and $p_i$ is the 3D coordinate, we design a Neighborhood Communication Block (NCB) to extract the local feature $F_L$ of $P_{in}$. Firstly, the KNN algorithm is utilized to obtain a neighborhood set $\mathcal{N}\left(p_{i}\right)=\left\{p_{i 1}, p_{i 2}, \ldots, p_{i k}\right\}$ of every point $p_{i}$, where k is the neighborhood size. We compute the geometric context of $p_{i}$ as $[p_{i}, p_{i}-p_{ik}]$, where $[\cdot,\cdot]$ denotes the concatenate operation, and the geometric context vector $C_{geo} \in \mathbb{R}^{N_{in} \times k \times 6}$ aggregates the geometric context of all points in $P_{in}$. Then, a simple MLP is utilized to extract the point-wise feature $F_{in}= \{f_i\}_{i=1}^{N_{in}}$, where $f_{i} \in \mathbb{R}^c$ is the corresponding feature of $p_{i}$, c is the dimension of $f_{i}$. Similarly, KNN algorithm is utilized to obtain a neighborhood set $\mathcal{N}\left(f_{i}\right)=\left\{f_{i 1}, f_{i 2}, \ldots, f_{i k}\right\}$ of every feature $f_{i}$. The feature context of $f_{i}$ is calculated by $[f_{i}, f_{i}-f_{ik}]$ and the feature context vector $C_{f} \in \mathbb{R}^{N_{in} \times k \times 2c}$ aggregates the feature context of all features in $F_{in}$. The local feature $F_L \in \mathbb{R}^{N_{in} \times C_{L}}$ of the sparse point cloud can be calculated by

\begin{displaymath}
    F_{L}=\max _{k}\left[\alpha\left(C_{\mathit{geo}}\right), \beta\left(C_{\mathit {f}}\right)\right],
\end{displaymath}

where $\alpha$ and $\beta$ are two independent non-linear function implemented by MLPs. Before doing feature enhancement, the fused global feature $F$ is replicated and concatenate with $F_L$. The attention mechanism is suitable for the self semantic communication of the point-wise feature. Thus, we employs Self-Attention Transformer (SAT) block for feature enhancement and the architecture of SAT is the same as the self-attention layer in~\cite{vaswani2017attention}. SAT integrates the information from different points of the point-wise feature by applying the multi-head self-attention with residual connection, which can be formulated as

\begin{displaymath}
\begin{split}
Z = \mathit{LayerNorm}(Q&+\mathit{MultiHead}(Q, K, V)), \\
Q = X W^{Q}, K &= X W^{K}, V = X W^{V},
\end{split}
\end{displaymath}

where $X$ is the input feature, $LayerNorm(\cdot)$ denotes the layer normalization operation, $MultiHead(\cdot)$ denotes the multi-head self-attetion layer, $W^{Q} \in \mathbb{R}^{2C_{L} \times C_{Q}}$, $W^{K} \in \mathbb{R}^{2C_{L} \times C_{K}}$ and $W^{V} \in \mathbb{R}^{2C_{L} \times C_{V}}$ are linear transformation matrices. Subsequently, a feed forward network is utilized to add non-linearity, thus the output of SAT can be formulated as
\begin{displaymath}
\mathit{SAT}(X) = Z+FFN(Z),
\end{displaymath}
After 3 consecutive SAT blocks, we get the enhanced local feature $F_{L}^{\prime} \in \mathbb{R}^{N_{in} \times C_{L}^{\prime}}$.

\noindent{{\bf Feature Expansion and Displacement Generation.}}
Note that during the local feature embedding and enhancement step, the number of points in point-wise feature $F_{L}^{\prime}$ remain unchanged. In order to achieve the goal of upsampling, we use 1D transpose convolution for feature expansion. This operation is called point splitting~\cite{xiang2021snowflakenet}, which can be intuitively interpreted as splitting each point of the point-wise feature into multiple points, thus increase the number of points. Then, the local feature $F_{L}$ is replicated and concatenated with the expanded feature. In this way, local feature serves as a guidance throughout the entire upsampling process, which is crucial to recovering local geometric details. Inspired by ViPC~\cite{zhang2021view} and CSDN~\cite{zhu2023csdn}, we also adpot the idea of predicting point-wise displacement for refinement. The displacement vector $\Delta P \in \mathbb{R}^{rN_{in} \times 3}$ is obtained by applying a shared MLP to the concatenated feature, where r is the upsample ratio. The output of SUT module can be formulated as

\begin{displaymath}
P_{\mathit{out}}=\mathit{replicate}\left(P_{\mathit{in}}, r\right)+\Delta P.
\end{displaymath}

\subsection{Loss Function}

Like the current state-of-the-art work~\cite{aiello2022cross}, we utilize L1 Chamfer Distance (L1-CD) as our loss function, which can be written as:
\begin{displaymath}
\mathcal{L}_{CD}(Y,{Y}_{gt})=\frac{1}{2|Y|}\sum_{y \in Y}\min_{\hat{y} \in {Y}_{gt}}||y - \hat{y}|| + \frac{1}{2|{Y}_{gt}|}\sum_{\hat{y} \in {Y}_{gt}}\min_{y \in Y}||\hat{y} - y||,
\end{displaymath}
where $Y$ is the predicted point cloud and ${Y}_{gt}$ is the ground truth. Chamfer Distance (CD) calculates the average closest point distance between the predicted point cloud and the ground truth point cloud, where the first term forces the predicted points to lie close to the ground truth points and the second term ensures the ground truth point cloud is covered by the predicted point cloud.

In order to explicitly constrain point clouds generated at each stage of our network, we calculate L1-CD for coarse point cloud $P_0$, the intermediate point cloud $P_1$ and the complete point cloud $P_C$. The corresponding ground-truth point clouds $Y_0$ and $Y_1$ are downsampled by the FPS algorithm from the ground truth $Y_{gt}$ to the same resolution as $P_0$ and $P_1$ respectively. The total training loss of our network is defined as
\begin{displaymath}
    \begin{split}
        \mathcal{L} = &\mathcal{L}_{CD}\left(P_0, Y_0\right) + \mathcal{L}_{CD}\left(P_1, Y_1\right) + 
        \mathcal{L}_{CD}\left(C, Y_{gt}\right).
    \end{split}
\end{displaymath}

\section{Experiments}

\subsection{Datasets}
The dataset used in our experiment is the benchmark dataset ShapeNet-ViPC~\cite{zhang2021view}, which is derived from ShapeNet~\cite{chang2015shapenet}. It contains 38,328 objects from 13 categories, including airplane, bench, cabinet, car, chair, monitor, lamp, speaker, firearm, sofa, table, cellphone and watercraft. For each object, there are 24 partial point clouds with occlusions generated under 24 viewpoints, which follow the same settings as ShapeNetRendering~\cite{chang2015shapenet}. The complete ground-truth point cloud is generated by uniformly sampling 2,048 points from the mesh surface of a target in ShapeNet~\cite{chang2015shapenet}. Moreover, each 3D shape is rotated to the pose corresponding to a specific view point and normalized into the bounding sphere with radius of 1. Images are generated from the same 24 viewpoints as ShapeNetRendering with a resolution of $224 \times 224$. Each object has 24 partial point clouds and 24 corresponding images rendered from 24 viewpoints and a ground-truth point cloud. During training, we randomly chose a viewpoint for the image and the partial point cloud. (The viewpoint of the two modalities can be different.) For the experiments conducted on the known categories, we employ the same experimental setting as ViPC~\cite{zhang2021view}, i.e., we use 31,560 objects from eight categories with $80 \%$ for training and $20\%$ for testing. And for the experiments conducted on novel categories, the training set remains unchanged and we use objects from 4 categories for testing which are not used in the training set.

\begin{table*}
    \captionsetup{justification=centering}
    \caption{Quantitative comparison for the point cloud completion on ShapeNet-ViPC dataset using per-point L2 Chamfer distance $\times 10^{-3}$ (lower is better). The best results are highlighted in bold.}
    \centering
    \begin{tabular}{c|c|cccccccc}
        \hline
        Methods & Avg & Airplane & Cabinet & Car & Chair & Lamp & Sofa & Table & Watercraft \\
        \hline
        \hline
        \multicolumn{10}{c}{Unimodal Methods} \\
        \hline
        AtlasNet~\cite{groueix2018papier} & 6.062 & 5.032 & 6.414 & 4.868 & 8.161 & 7.182 & 6.023 & 6.561 & 4.261 \\
        FoldingNet~\cite{yang2018foldingnet} & 6.271 & 5.242 & 6.958 & 5.307 & 8.823 & 6.504 & 6.368 & 7.080 & 3.882 \\
        PCN~\cite{yuan2018pcn} & 5.619 & 4.246 & 6.409 & 4.840 & 7.441 & 6.331 & 5.668 & 6.508 & 3.510 \\
        TopNet~\cite{tchapmi2019topnet} & 4.976 & 3.710 & 5.629 & 4.530 & 6.391 & 5.547 & 5.281 & 5.381 & 3.350 \\
        ECG~\cite{pan2020ecg} & 4.957 & 2.952 & 6.721 & 5.243 & 5.867 & 4.602 & 6.813 & 4.332 & 3.127 \\
        VRC-Net~\cite{pan2021variational} & 4.598 & 2.813 & 6.108 & 4.932 & 5.342 & 4.103 & 6.614 & 3.953 & 2.925 \\
        PointAttn~\cite{wang2024pointattn} & 2.853 & 1.613 & 3.969 & 3.257 & 3.157 & 3.058 & 3.406 & 2.787 & 1.872 \\
        \hline
        \multicolumn{10}{c}{Multimodal Methods} \\
        \hline
        ViPC~\cite{zhang2021view} & 3.308 & 1.760 & 4.558 & 3.183 & 2.476 & 2.867 & 4.481 & 4.990 & 2.197 \\
        CSDN~\cite{zhu2023csdn} & 2.570 & 1.251 & 3.670 & 2.977 & 2.835 & 2.554 & 3.240 & 2.575 & 1.742 \\
        XMFnet~\cite{aiello2022cross} & 1.443 & 0.572 & 1.980 & 1.754 & 1.403 & 1.810 & 1.702 & 1.386 & 0.945 \\
        \hline
        Ours & \bf{1.038} & \bf{0.453} & \bf{1.678} & \bf{1.462} & \bf{1.019} & \bf{0.583} & \bf{1.340} & \bf{1.072} & \bf{0.696} \\
        \hline
    \end{tabular}
    \label{tab:table1}
\end{table*}

\subsection{Implementation Details and Evaluation Metrics}
In our implementation, the input partial point cloud $P$ contains $N = 2048$ points. Follow XMFnet~\cite{aiello2022cross}, the EdgeConv layers chose $k = 20$ as neighborhood size and the SAGPool layers chose $k = 16$ and $k = 6$ as neighborhood size. Each SAGPool layer downsamples the partial point cloud by a factor of 4. The output feature size of the 3D encoder is $2,048/16 \times 512 = 128 \times 512$. The output feature map of the 2D encoder has a size of $7 \times 7 \times 512$, and by reshaping, the size of the pixel-wise feature is $49 \times 512$. The output feature dimensions of the DMF module is 512. The coarse point cloud $P_0$ contains $N_0 = 256$ points and the partial point cloud $P$ is downsampled by the FPS algorithm to 256 points and concatenated with $P_0$ leading to a point cloud with 512 points. The NCB chose $k = 16$ as neighborhood size. The dimension of the local feature $F_L$ is $C_L = 128$. The dimension of the enhanced local feature $F_{L}^{\prime}$ is $C_{L}^{\prime} = 512$. Each SUT module has the upsample ratio of 2, which means the output point cloud of the first SUT module contains $N_1 = 512 \times 2 = 1024$ points and the complete point cloud contains $N_C = 1024 \times 2 = 2048$ points.

For quantitative evaluation, like previous works, we chose L2 Chamfer Distance (L2-CD) and F-Score as evaluation metrics. L2-CD can be formulated as

\begin{displaymath}
\mathcal{L}_{CD}(Y,{Y}_{gt})=\frac{1}{2|Y|}\sum_{y \in Y}\min_{\hat{y} \in {Y}_{gt}}||y - \hat{y}||_2^2 + \frac{1}{2|{Y}_{gt}|}\sum_{\hat{y} \in {Y}_{gt}}\min_{y \in Y}||\hat{y} - y||_2^2,
\end{displaymath}

The whole network is trained end-to-end with an initial learning rate of $1 \times 10^{-4}$ for 120 epochs with a batch size of 16. Adam optimizer~\cite{kingma2014adam} is utilized during the training process with $\beta_1 = 0.9$ and $\beta_2 = 0.999$. The learning rate is decayed by 0.7 for every 20 epochs. The proposed network is implemented by Pytorch and trained on NVIDIA RTX 3090 GPU.

\subsection{Experiment Results on Known Categories}
In this section, we will compare our DMF-Net with several existing point cloud completion networks. We compare the results of several unimodal methods which only take the partial point cloud as input, including AtlasNet~\cite{groueix2018papier}, FoldingNet~\cite{yang2018foldingnet}, PCN\cite{yuan2018pcn}, TopNet~\cite{tchapmi2019topnet}, ECG\cite{pan2020ecg}, VRC-Net~\cite{pan2021variational} and PointAttN~\cite{wang2024pointattn}. AtlasNet and FoldingNet generate the complete point cloud by using 2D grids for folding operation. PCN employs an encoder-decoder architeture to recover the 3D shape in a coarse-to-fine manner. TopNet generates structured point clouds with a tree-like decoder. ECG emloys an auto-encoder to recover the skeleton of a complete point cloud and applies convolution on edges for refinement. VRC-Net proposes a probabilistic modeling and relational enhancement network for coarse-to-fine point cloud completion. PointAttN proposes a coarse-to-fine completion network with attention mechanism throughout the whole completion process. We also compare the results of existing multi-modal point cloud completion methods, including ViPC~\cite{zhang2021view}, XMFnet~\cite{aiello2022cross} and CSDN~\cite{zhu2023csdn}. ViPC is the pioneering work for leveraging a single-view image for point cloud completion. XMFnet adpots cross- and self-attention layers for modality fusion and proposes a decoder with several independent braches for generating different regions of the complete point cloud. CSDN reformulate the modality fusion problem as a style transfer problem and predicts offset vectors for refinement. For the above methods, the output predicted complete point cloud contains 2,048 points.

\noindent{{\bf Quantitative Results.}}
Follow XMFnet~\cite{aiello2022cross}, we calculated CD on the 2,048 points of each object. A lower CD value indicates better performance. Category-specific training is performed and the results for each category is reported in Table~\ref{tab:table1}. It can be seen that the performance of multimodal completion networks is generally better than that of unimodal completion networks, proving that the introduction of image modality is effective for point cloud completion. The proposed DMF-Net achieves the best results across all 8 categories in terms of CD. It is worth noting that compared with the current top-ranked XMFnet, DMF-Net reduces the average CD value by 0.405, which is $28.1\%$ lower than the results of XMFnet. Moreover, each category has different degrees of improvement in completion results, with a significant improvement in the lamp category, proving that DMF-Net can handle objects with slender structures well.

\noindent{{\bf Qualitative Results.}}
Fig.~\ref{fig:results} shows the point clouds completed by the proposed DMF-Net and its competitors, including two unimodal methods (PCN~\cite{yuan2018pcn} and PointAttN~\cite{wang2024pointattn}) and two multimodal methods (ViPC~\cite{zhang2021view} and XMFnet~\cite{aiello2022cross}). Take the airplane category as an example, the partial point cloud has a large degree of incompleteness in its right wing and right tail wing. PCN can restore the global shape of the airplane, but local details are severely distorted. PointAttN and ViPC perform slightly better in local detail restoration than PCN, but generate noisy points near the surface. XMFnet produce relatively less noise while recovering local details, but it can be clearly observed that the missing right wing and right tail are relatively sparse, and the distribution of points is nonuniform. Our DMF-Net not only restores fine local geometric structures, but also ensures the uniformity of points. Besides, the overall shapes of complete point clouds generated by the proposed DMF-Net are cloest to the ground-truth.

\section{Discussion}
\begin{figure*}[htbp]
    \centering
    \includegraphics[width=\textwidth]{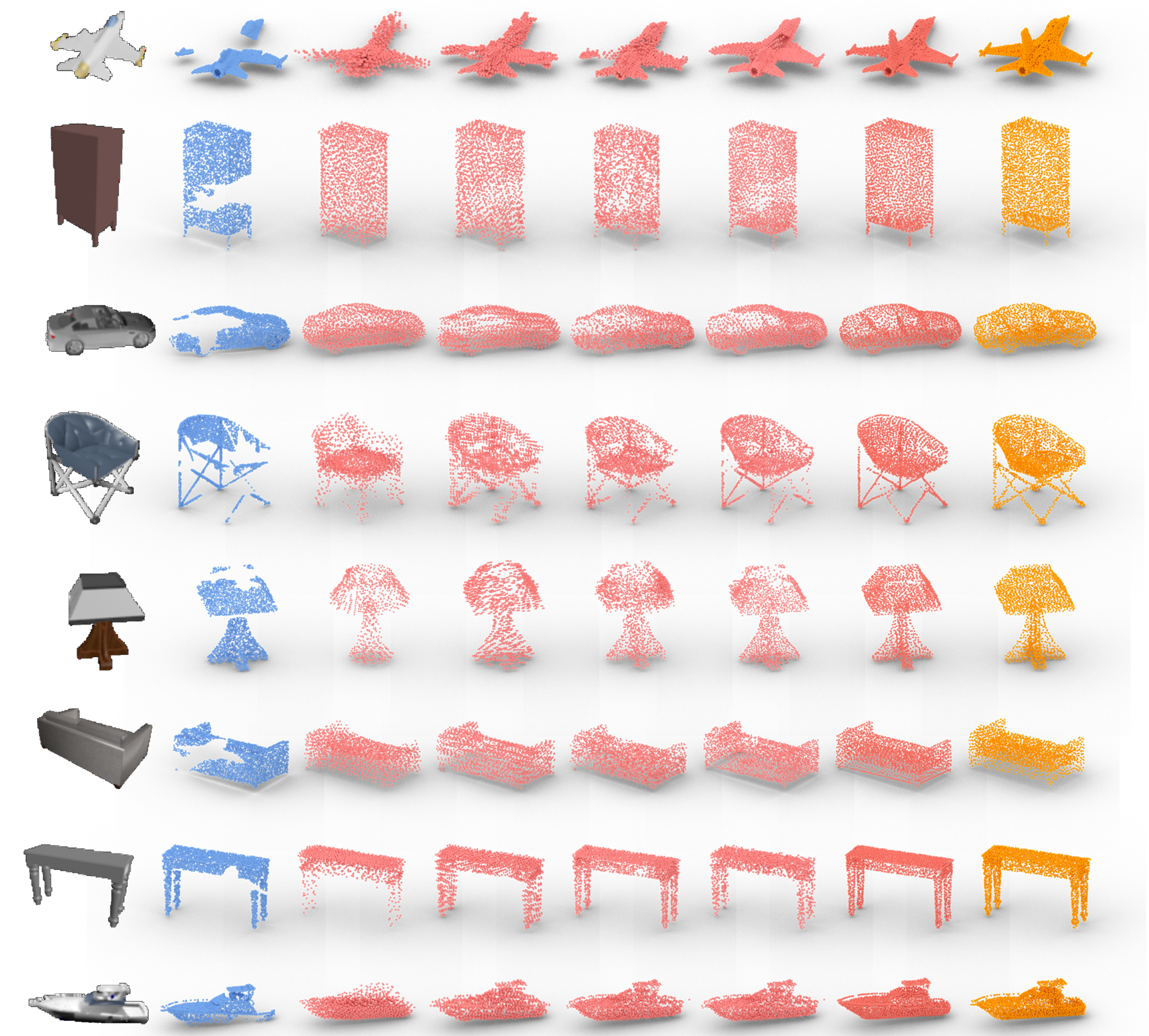}
    \begin{tabularx}{0.95\textwidth}{@{}Y@{}Y@{}Y@{}Y@{}Y@{}Y@{}Y@{}Y@{}}
    Image \ \ \ &
    Partial \ \ \ &
    PCN \ \ \ &
    PointAttN \ \ \ \ &
    ViPC \ \ \ \ \ &
    XMFnet \ \ \ \ \ &
    Ours \ \ \ \ \ \ \ &
    GT \ \ \ \ \ \ 
    \end{tabularx}
    \caption{Qualitative completion results produced by PCN~\cite{yuan2018pcn}, PointAttN~\cite{wang2024pointattn}, ViPC~\cite{zhang2021view}, XMFnet~\cite{aiello2022cross} and DMF-Net on known categories of the ShapeNet-ViPC dataset. Our DMF-Net produces the most complete and uniform point cloud with local details and compared to its competitors.}
    \Description{Qualitative results on known categories.}
    \label{fig:results}
\end{figure*}

In this paper, we propose a novel image-guided point cloud completion network with dual-channel modality fusion and shape-aware upsampling transformer, which can recover an uniform complete point cloud with rich local details in a coarse-to-fine manner. However, there are still limitations to the method. For instance, as shown in Fig.~\ref{fig:results}, the fourth row shows the completion results on a folding chair. Although our DMF-Net recovers the backrest and the seat of the chair, the completion results of the four legs are less satisfied. This is largely due to the detailed structures (i.e., the legs of the chair) of the partial point cloud are relatively sparse. Meanwhile, DMF-Net relies heavily on the supervision of the ground truth point cloud which is hard to acquire in the real-world scenarios. Besides, the 
majority of the input images are of very low resolutions, which means the complementary information provided by these images is very limited. To tackle this, utilizing a third modality (e.g. text) is a common practice. Inspired by the vision-language model CLIP~\cite{radford2021learning}, ULIP~\cite{xue2023ulip} proposes a unified framework that can improve 3D understand by aligning features from point cloud, image and text in the same space through contrastive learning. Song et al.~\cite{song2023fine} propose a fine-grained text corpus generated by pre-trained large language models for text and image guided point cloud completion task. These works have shown us the possibility of introducing text modalities into point cloud completion task which can provide more complementary information. By making slight modifications to our overall framework, we have made preliminary attempt on the utilization of text modality. Specifically, inspired by ULIP, we adopt template-based descriptions as the text input. Then, through the pre-trained CLIP model, we get the features of text modality and image modality. During the modality fusion process, we continue to use the idea of symmetric fusion while the decoder and upsampler remain unchanged. However, due to the length limitations, the implementation details and the 
experimental results of the mentioned extension method are not presented in this paper.

Our future work can be conducted in the following aspects. Firstly, exploring the generation method of fine-grained text (e.g. descriptions of global symmetry and local detail shapes) which is capable of providing more complementary information.
Secondly, employing fine-tuning techniques on the pre-trained network to make them more suitable for point cloud completion task. Finally, how to improve the performance of point cloud completion with the help of multi-view images is also worth exploring.

\clearpage


\bibliographystyle{ACM-Reference-Format}
\bibliography{sample-base}

\end{document}